\documentclass[lettersize,journal]{IEEEtran}
\usepackage{amsmath,amsfonts}
\usepackage{algorithmic}
\usepackage{algorithm}
\usepackage{array}
\usepackage[caption=false,font=normalsize,labelfont=sf,textfont=sf]{subfig}
\usepackage{textcomp}
\usepackage{stfloats}
\usepackage{url}
\usepackage{verbatim}
\usepackage{graphicx}
\usepackage{cite}
\usepackage{booktabs}
\usepackage{multirow}

\usepackage{makecell}
\usepackage{pifont}
\usepackage{enumitem}
\usepackage{xcolor}

\usepackage{marvosym}

\newcommand{\cmark}{\ding{51}}%
\newcommand{\xmark}{\ding{55}}%

\begin{document}

\title{TimeSeek: Multi-Task and Multi-Domain Pretraining for
Generalized Time Series Analytics}

\author{IEEE Publication Technology,~\IEEEmembership{Staff,~IEEE,}
\thanks{This paper was produced by the IEEE Publication Technology Group. They are in Piscataway, NJ.}
\thanks{Manuscript received April 19, 2021; revised August 16, 2021.}}

\author{Peng~Chen,
        Yihang~Wang,
        Zhongwen~Rao, 
        Yang~Shu,
        Yuying~Qiu,
        Yuxuan~Chen,
        Lujia~Pan,\\
        Bin~Yang, ~\IEEEmembership{Senior Member,~IEEE,}
        Chenjuan~Guo\textsuperscript{\Letter}
\IEEEcompsocitemizethanks{
\IEEEcompsocthanksitem Peng Chen, Yihang Wang, Yang Shu, Yuying Qiu, Yuxuan Chen, Bin Yang, and Chenjuan Guo are with the School of Data Science and Engineering,  East China Normal University, Shanghai 200333, China. (email: \{pchen, yhwang, yyqiu, chenyx\_50\}@stu.ecnu.edu.cn, \{yshu,byang,cjguo\}@dase.ecnu.edu.cn). 
\IEEEcompsocthanksitem Zhongwen Rao and Lujia Pan are with the Noah’s Ark Lab, Hua Wei, China. (email:\{raozhongwen, panlujia\}@huawei.com)
\IEEEcompsocthanksitem Chenjuan Guo is the corresponding author. 
}
}

\markboth{}{}%


\maketitle

\begin{abstract}
Time series analytics has been widely applied across various domains. While existing time series foundation models (TSFMs) have demonstrated promising results, they still face two major challenges:  1) Most TSFMs mainly rely on stacking Transformer layers to increase parameter size, lacking the ability to adaptively model complex time series patterns from multi-domain data. 2) Current TSFMs are primarily pre-trained for specific tasks, e.g., forecasting or anomaly detection, and a few models support multiple tasks, but they fail to fully leverage task specificity and commonality, resulting in suboptimal performance. To address these challenges, we propose TimeSeek, a novel generalized time series analytics model that adaptively models time series patterns from multi-domain data and unifies multiple tasks in pre-training. 
Conceptually, TimeSeek is guided by selective information sharing under domain and task heterogeneity. From the perspective of multi-domain data, we propose a Multi-Domain Feature Extractor that adaptively captures temporal dependencies and channel correlations based on domain-specific characteristics, including progressive multi-scale modeling and adaptive slot channel modeling. From the perspective of the diversity of tasks, we propose multi-task collaborative training, which leverages task specificity and task commonality through task customization and cross-task training, effectively unifying multiple tasks in pre-training and better adapting to diverse downstream scenarios. After pre-training on large-scale time series data, TimeSeek achieves state-of-the-art performance in forecasting, anomaly detection, and classification tasks across both few-shot learning and zero-shot situations, together with favorable inference efficiency. 
\end{abstract}


\section{Introduction}

 \begin{figure}[h]  
    \centerline{\includegraphics[width=0.9\linewidth]{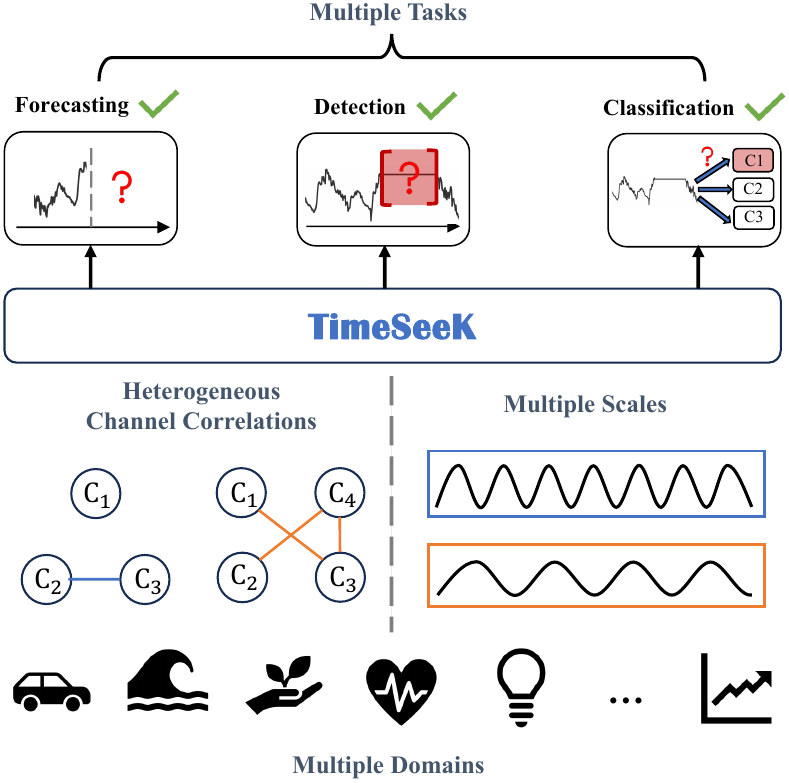}}\vspace{-2mm}
    \caption{TimeSeek leverages multi-domain data and unifies multiple tasks in pre-training.}
    \vspace{-1em}
    \label{fig: intro}
\end{figure}
\IEEEPARstart{W}{ith} the rapid advancement of the Internet of Things, a vast amount of multi-domain time series data has been generated in real-world scenarios, leading to increasing demand for time series analytics, including time series forecasting, anomaly detection, and classification tasks. Time series analytics primarily focuses on modeling complex time series patterns to uncover underlying regularities, thereby providing robust support for decision-making \cite{designing}. It has been widely applied across various domains such as finance \cite{master}, meteorology \cite{airformer}, transportation \cite{use}, and energy \cite{gas}.

In recent years, Foundation Models have emerged as a groundbreaking paradigm in deep learning, demonstrating unprecedented success in fields like natural language processing \cite{llama, gpt3} and computer vision \cite{ViT,swintransformer}. These models leverage massive datasets and large-scale parameter architectures to learn unified representations, exhibiting exceptional generalization capabilities and adaptability across different tasks. Inspired by these achievements, researchers have begun exploring the potential of foundation models for the time series field \cite{timer, moment, timemoe,timesfm}.

Existing Time Series Foundation Models (TSFMs) \cite{timer, moment, timesfm} typically adopt a two-stage paradigm: pre-training on multi-source time series data followed by fine-tuning or direct inference for downstream scenarios. As shown in Figure \ref{fig: intro}, time series data exhibit distinct temporal dependencies and channel correlations for different domains. In the temporal aspect, time series data often display hierarchical temporal patterns—spanning daily, weekly, and seasonal cycles—with the significance of each scale varying considerably across domains. In the channel aspect, channel correlations show significant inter-domain variability. In some domains, such as IoT sensor networks, strong channel dependencies are inherently present, while in others, such as medical vital sign monitoring, correlations may be comparatively weaker. Therefore, effectively modeling multi-scale features and channel correlations during multi-domain pre-training is necessary.

Furthermore, downstream scenarios involve a variety of tasks, including forecasting, anomaly detection, and classification. The prevalent task-specific pre-training approaches in  TSFMs \cite{timemoe, timesfm, moirai} not only incur substantial computational costs due to every task requiring a specific foundation model but also limit the model’s capacity to a single task. In contrast, establishing a unified pre-training paradigm that harmonizes heterogeneous tasks would enable more effective adaptation to multi-task scenarios. Considering both the heterogeneity of domain data and the diversity of tasks, we identify two key challenges that need to be addressed to advance time series foundation models.

\textit{For the heterogeneity of domain data}, time series data across different domains exhibit distinct multi-scale and channel correlation patterns. Adaptively modeling these patterns can significantly enhance performance, yet existing methods fall short here. For multi-scale modeling, current TSFMs \cite{moirai, timer, timesfm, units} are typically designed for a single scale, which limits their ability to capture multi-scale features required across different domains. While models like MOIRAI \cite{moirai} adapt patch sizes for different frequency domains, they still lack dynamic scale integration.
For correlation modeling, most TSFMs \cite{timer, timesfm, moment, chronos, timemoe} adopt a channel-independent approach, overlooking the intrinsic correlations between channels. Moreover, the few TSFMs \cite{moirai} that employ a uniform channel modeling strategy across all domains have the risk of introducing spurious correlations among inherently uncorrelated channels, leading to suboptimal generalization. 
Therefore, a flexible modeling approach that can dynamically capture two patterns for each domain is essential.

\textit{For the diversity of tasks}, time series foundation models are expected to serve as versatile tools for diverse downstream applications, necessitating the capability for multi-task learning. While time series tasks share some commonalities, they vary significantly in their objectives and requirements. This inherent heterogeneity can complicate the balancing of learning objectives during pre-training and may even lead to a negative impact between tasks.
Existing TSFMs \cite{timemoe, timesfm, moirai} are primarily pre-trained for specific tasks, which not only incur substantial computational costs due to every task requiring a specific foundation model but also restrict the model’s ability to generalize across tasks. While a few TSFMs support multiple tasks, they fail to balance task specificity and commonality. For instance, Moment \cite{moment} relies on reconstruction pretraining and requires swapping specific task heads for fine-tuning, while UniTS \cite{units} simply utilizes task tokens without addressing task heterogeneity.
Therefore, designing a pre-training framework that unifies task commonality and specificity to improve overall task performance is necessary.

To address these challenges, we propose Multi-Task and Multi-Domain Pre-training for Generalized Time Series Analytics, namely \textbf{TimeSeek}, to adaptively model time series patterns of multi-domain data and unify multiple time series tasks in pre-training. Conceptually, TimeSeek is motivated by selective information sharing under domain and task heterogeneity. Instead of assuming that information should be uniformly exchanged across all temporal scales, channels, and tasks, it learns when and how such sharing is beneficial while retaining task- and domain-specific representations.

\textbf{For the first aspect:} To better model time series patterns for multi-domain data, we propose a Multi-Domain Feature Extractor that captures complex temporal dependencies and channel correlations. 
\textit{In the temporal aspect}, we propose progressive multi-scale modeling. It gradually fuses features from coarse-grained to fine-grained features and incorporates a gating mechanism to dynamically adjust the weights of each scale. Through this approach, the model adaptively filters redundant features and highlights key scale information, enhancing its generalization and adaptability.
\textit{In the channel aspect}, we propose adaptive slot channel modeling to capture channel correlations in multi-domain data. It consists of two steps: 1) the channel router adaptively determines whether meaningful channel correlations exist in the input; if so, 2) channel slot attention adaptively groups the strongly correlated channels into the same slots, combined with a masking strategy to dynamically filter less informative slots and retain important ones. 
Together, these designs selectively transfer information across temporal scales and channel dependencies, enabling TimeSeek to adapt to heterogeneous multi-domain data.

\textbf{For the second aspect:} To unify the multi-task time series pre-training framework and leverage task specificity and commonality, we propose Multi-Task Collaborative Training. It incorporates task-customized pre-training modules alongside cross-task collaborative training. \textit{In task customization}, we design specialized modules for three core time series tasks: a non-autoregressive predictor for forecasting, a patch-based reconstruction module for anomaly detection, and a global feature extractor for classification. These innovative designs equip TimeSeek with capabilities tailored to specific tasks, such as generative and discriminative abilities. \textit{In the cross-task training}, 
these customized modules are structurally unified over the shared multi-domain feature extractor and collaboratively optimized via a dual-loss mechanism. Crucially, tasks mutually reinforce each other; For example, the global prior extracted from the classification task enhances the semantic consistency of anomaly detection, while the gradients propagated from the reconstruction loss simultaneously improve the classifier's feature extraction capabilities. Specifically, our contributions can be summarized as follows:

\begin{itemize}
    \item We propose TimeSeek, a novel generalized time series analytics model that adaptively models time series patterns from multi-domain data and unifies multiple tasks in pre-training, including forecasting, anomaly detection, and classification tasks.

    \item We propose a multi-domain feature extractor, including progressive multi-scale modeling and adaptive slot channel modeling to effectively capture temporal dependencies and channel correlations from multi-domain data.

    \item We propose multi-task collaborative training to effectively unify the multi-task pre-training and balance task specificity and commonality.

    \item After pre-training on large-scale time series data, TimeSeek achieves state-of-the-art performance on three time series tasks across both few-shot learning and zero-shot inference situations.
\end{itemize}

\section{Related Work}
Time series analytics has garnered extensive research attention~\cite{Timesnet}, extending to several time series tasks such as time series forecasting~\cite{tfb, faloutsos2018forecasting,musebench}, time series anomaly detection~\cite{anomly_survey, liu2024time, tsb, aus_roc}, and time series classification~\cite{classification_survey}. Existing models can be categorized into two types: time-series-specific models, which are trained and tested on each specific dataset, and time series foundation models, which leverage pre-training on multi-domain datasets and evaluate on downstream datasets.

\subsection{Time-Series-Specific Models}
Although the time-series-specific models can achieve good results on their respective tasks, they all need to be retrained and require tailored parameter tuning for each specific dataset, which causes high costs and limits their performance in data-scarce scenarios.

For the task of time series forecasting, with the development of deep learning, many works based on MLP~\cite{dlinear, fits, zhong2024multi}, RNN~\cite{deepar, MileTS}, CNN~\cite{Timesnet, enhancenet}, and GNN~\cite{zhao2023multiple, autocts, metro, ma2024learning} emerged.
DLinear~\cite{dlinear} decomposes a raw data input into a trend component and a seasonal component and then applies two one-layer linear layers to each component. 
TimesNet~\cite{Timesnet} models multi-period temporal variations through adaptively aggregated two-dimensional representations.
In recent years, the Transformer has shown great advantages in capturing long-term dependencies~\cite{triformer, crossformer, patchtst, pathformer, iTransformer}. 
PatchTST~\cite{patchtst} segments time series into subseries-level patches that serve as input tokens to the Transformer and applies the channel-independence strategy for training on multivariate time series. 
CrossFormer~\cite{crossformer} captures hierarchical cross-time and cross-dimension dependencies, and iTransformer~\cite{iTransformer} learns cross-variable correlations through attention over variate tokens. These methods provide multi-scale or multivariate representations in task-specific training settings. TimeSeek addresses the additional challenge of selectively controlling information sharing across temporal scales, channels, and task-specific pathways within multi-domain and multi-task pre-training.

For the task of time series classification ~\cite{classification_survey}, 
deep learning-based methods~\cite{Classification_cnn, LSTM-FCNs, tsec, inceptiontime, lite, convtran} require less domain knowledge from time series data to handle classification tasks than traditional methods~\cite{UFS, gRSF, shapelet}, but these methods require a large amount of labeled data to train model parameters.
To address this, some works have explored prototype-based networks~\cite{tapnet} or representation learning~\cite{ts2vec, ts-tcc, cost, simMTM}, enhancing classification performance.

For the task of time series anomaly detection, 
deep learning-based methods~\cite{DAGMM,LSTM-NDT, OmmiAnomaly,moderntcn, daemon} mainly include reconstruction-based methods and prediction-based methods.
Reconstruction-based methods~\cite{ae1,vae1, robust, fang2024temporal, D3R, anomaly} compress the original input data and then attempt to reconstruct the series, treating high reconstruction errors as anomalies.
Prediction-based methods~\cite{LSTM-NDT} utilize past observations to predict current values and use the difference between the predicted and true values as an anomaly criterion.

\subsection{Time Series Foundation Models}
Recently, the foundation model for time series has gained significant attention~\cite{timer, chronos, moirai, timesfm, moment, TTMs, rose, LPTM, timemoe, horai}. 
By leveraging pre-training on large-scale and diverse time series datasets, time series foundation models demonstrate remarkable adaptability to new tasks with minimal fine-tuning, while even achieving zero-shot transfer capabilities across domains.
Timer~\cite{timer} adopts a decoder-only architecture employing autoregressive modeling for generative pre-training.
Chronos~\cite{chronos} converts time series data into discrete tokens by scaling and quantization with a fixed vocabulary, then trains a Transformer-based language architecture on these tokenized time series.
TiRex \cite{tirex} targets zero-shot forecasting with an xLSTM-based architecture, FlowState \cite{flowstate} focuses on sampling-rate-invariant forecasting, and Chronos-2 \cite{chronos-2} extends zero-shot forecasting to univariate, multivariate, and covariate-informed settings.
Beyond the autoregressive paradigm, other generative paradigms have been explored to enhance time series foundation models. TEMPO~\cite{tempo} adapts pre-trained language models by decomposing time series into trend, seasonal, and residual components, utilizing prompt-based learning to facilitate distribution adaptation across non-stationary domains. TimeDiT~\cite{timedit} introduces a Diffusion Transformer architecture equipped with unified masking schemes, coupling transformer-based temporal dependency learning with diffusion processes to capture uncertainties and generate high-quality probabilistic forecasts.
However, the above approaches mainly focus on the problem of time series forecasting, and it is difficult to take into account other tasks.

Research on time series foundation models that unify the three tasks of time series forecasting, classification, and anomaly detection remains at a nascent stage, but has shown great potential.
Moment~\cite{moment} pre-trains via masked reconstruction with a lightweight decoder, but requires task-specific heads in downstream forecasting scenarios, limiting its zero-shot ability.
UNITS~\cite{units} introduces task tokens to unify generative tasks and discriminative tasks into one format. However, it does not fully consider the balance between task heterogeneity and commonality in pre-training.

\textbf{Summary:} Table~\ref{tsf categorize} compares representative TSFMs in terms of multi-task pre-training, multi-scale modeling, channel modeling, and zero-shot inference. TimeSeek's distinction lies in selectively sharing information across temporal scales, channel dependencies, and task-specific pathways under domain and task heterogeneity.

\begin{table}[htbp]
  \centering
  \caption{Compare TimeSeek with existing TSFMs.}
  \resizebox{1\linewidth}{!}{
      \begin{tabular}{c c c c c}
      \toprule
      \textbf{Models} & \textbf{\makecell{Multi-Task \\ Pre-training}} & \textbf{\makecell{Multi-Scale \\ Modeling}} & \textbf{\makecell{Channel \\ Modeling}} & \textbf{Zero-Shot} \\
      \midrule
      Timer \cite{timer} & \xmark & \xmark & \xmark & \cmark \\
      MOIRAI \cite{moirai} & \xmark & \xmark & \cmark & \cmark \\
      TimesFM \cite{timesfm} & \xmark & \xmark & \xmark & \cmark \\
      Chronos \cite{chronos} &\xmark & \xmark & \xmark & \cmark \\
      Moment \cite{moment} &\xmark & \xmark & \xmark & \xmark \\
      UniTS \cite{units} & \cmark & \xmark & \cmark & \cmark \\
      Time-MOE \cite{timemoe} & \xmark & \xmark & \xmark & \cmark \\
      TEMPO \cite{tempo} & \xmark & \xmark & \xmark & \cmark \\
      TimeDiT \cite{timedit} & \xmark & \xmark & \xmark & \cmark \\
      \textbf{TimeSeek (our model)} & \cmark & \cmark & \cmark & \cmark \\
       \bottomrule
      \end{tabular}
    }
  \label{tsf categorize}%
  \vspace{-1em}
\end{table}%

\section{Preliminary}
\subsection{Time Series}
Time series exhibit the fluctuation of observed data across time. Given a series of length $L$, denoted by $ \mathbf{X} = \langle \mathbf{x}_1, \mathbf{x}_2, \ldots, \mathbf{x}_L \rangle $, every $ \mathbf{x}_t \in \mathbb{R}^C $ represents the values from $C$ different channels at the timestampe $t$. When there is more than one channel ($C > 1 $), the series is termed multivariate; conversely, if there is only a single channel ($ C = 1 $), it is referred to as univariate.

\subsection{Time Series Analytics}
Time series analytics studies historical time series data to understand the underlying patterns, supporting various tasks and decision-making. In this paper, we study time series forecasting, anomaly detection, and classification tasks.

\textbf{Time Series Forecasting} refers to the task of predicting future values based on past observations. Given a sequence of inputs \( \langle \mathbf{x}_{t-T+1}, \mathbf{x}_{t-T+2}, \dots, \mathbf{x}_{t} \rangle \), the goal is to forecast the future sequence \( \langle \mathbf{x}_{t+1}, \mathbf{x}_{t+2}, \dots, \mathbf{x}_{t+F} \rangle \), where \( T \)  and $F$ are the length of the input sequence and the real future sequence. This forecasting problem can be formalized as a mapping function \( \mathcal{F} \), as shown below:
\begin{equation}
    \begin{aligned}
        \mathcal{F}_{\theta}(\mathbf{x}_{t-T+1}, \mathbf{x}_{t-T+2}, \dots, \mathbf{x}_{t}) = (\mathbf{\hat{x}}_{t+1}, \mathbf{\hat{x}_{t+2}}, \dots, \mathbf{\hat{x}_{t+F}}),
    \end{aligned}
\end{equation}
where \( \theta \) refers to the learnable parameters of the model \( \mathcal{F} \), and \( \langle \mathbf{\hat{x}}_{t+1}, \mathbf{\hat{x}}_{t+2}, \dots, \mathbf{\hat{x}}_{t+F} \rangle \) represents the predicted future values.

\textbf{Time Series Anomaly Detection} focuses on identifying points or intervals in a time series that significantly deviate from normal patterns. Given a sequence of inputs \( \langle \mathbf{x}_{1}, \mathbf{x}_{2}, \dots, \mathbf{x}_{T} \rangle \), the goal is to detect whether the sequence contains any anomalies, represented as \( \langle y_1, y_2, \dots, y_T \rangle \), where \( y_t \in \{0, 1\} \) indicates whether the observation \( \mathbf{x}_t \) is anomalous. The anomaly detection problem can be formalized as a mapping function \( \mathcal{A} \), as shown below:
\begin{equation}
    \begin{aligned}
        \mathcal{A}_{\theta}(\mathbf{x}_{1}, \mathbf{x}_{2}, \dots, \mathbf{x}_{T}) = (y_1, y_2, \dots, y_T),
    \end{aligned}
\end{equation}
where \( \theta \) refers to the learnable parameters of the model \( \mathcal{A} \).

\textbf{Time Series Classification} aims to assign time series to predefined categories. Given a sequence of inputs \( \langle \mathbf{x}_{1}, \mathbf{x}_{2}, \dots, \mathbf{x}_{T} \rangle \), the objective is to classify it into the correct category \( y \in \{1, 2, \dots, M\} \), where \( M\) denotes the total number of classes. This classification task can be formulated as a mapping function \( \mathcal{C} \), as shown below:
\begin{equation}
    \begin{aligned}
        \mathcal{C}_{\theta}(\mathbf{x}_{1}, \mathbf{x}_{2}, \dots, \mathbf{x}_{T}) = y,
    \end{aligned}
\end{equation}
where \( \theta \) refers to the learnable parameters of the model \( \mathcal{C} \).

\subsection{Time Series Foundation Model}
Time series foundation model is pre-trained using a multi-source dataset \( \mathbf{D}_\mathrm{pre-train} = \{(X_t^j, Y_t^j)\}_{j=1}^N \), where \( N \) represents the number of pre-training datasets. For downstream tasks, the model is fine-tuned using the training set of the downstream dataset, \( \mathbf{D}_\mathrm{train} = \{\mathbf{X}_t^{\mathrm{train}}, \mathbf{Y}_t^{\mathrm{train}}\} \), and evaluated on the test set \( \mathbf{D}_\mathrm{test} = \{\mathbf{X}_t^{\mathrm{test}}, \mathbf{Y}_t^{\mathrm{test}}\} \) to infer \( \hat{\mathbf{Y}}_\mathrm{test} \). The datasets \( \mathbf{D}_\mathrm{pre-train} \), \( \mathbf{D}_\mathrm{train} \), and \( \mathbf{D}_\mathrm{test} \) are mutually exclusive. Alternatively, the model can directly perform inference on \( \mathbf{D}_\mathrm{test} \) without fine-tuning on \( \mathbf{D}_\mathrm{train} \), allowing it to predict \( \hat{\mathbf{Y}}_\mathrm{test} \) in a zero-shot manner. When applied to different tasks, the model can be viewed as \( \mathcal{F} \) for forecasting, \( \mathcal{A} \) for anomaly detection, and \( \mathcal{C} \) for classification.

\section{Methodology}
\subsection{Framework of TimeSeek}
 \begin{figure}[ht]  
    \centerline{\includegraphics[width=1\linewidth]{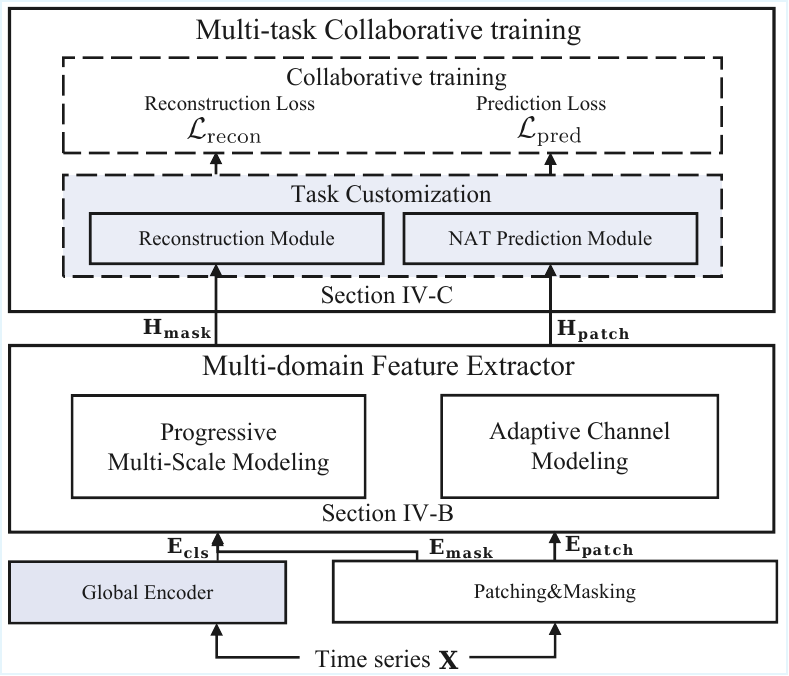}}
    \caption{The framework of TimeSeek.}
    \label{fig: timeseek_overview}
\end{figure}

 \begin{figure*}[ht]  
    \centerline{\includegraphics[width=1\linewidth]{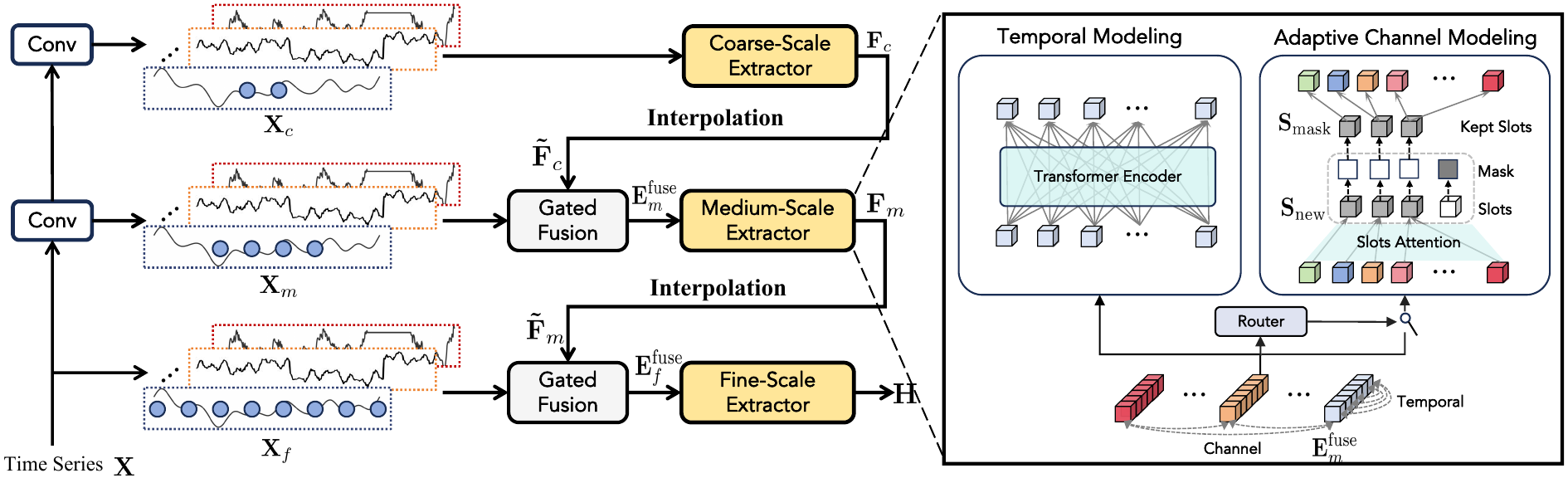}}
    \caption{The framework of the Multi-Domain Feature Extractor (MFE).}
    \label{fig: multi-domain feature extractor}
\end{figure*}

To build a versatile foundation model capable of handling diverse downstream applications, TimeSeek tackles two primary challenges: the heterogeneity of multi-domain data and the diversity of time series tasks. As illustrated in Figure \ref{fig: timeseek_overview}, the overall architecture introduces two core components:

\textbf{Multi-Domain Feature Extractor (MFE):} 
To address multi-domain data heterogeneity, the MFE serves as the shared, generalized backbone of TimeSeek. Unlike existing methods that rely on fixed-scale windows or channel-independent assumptions, the MFE adaptively captures complex temporal dependencies and channel correlations based on domain-specific characteristics. It achieves this through a progressive multi-scale modeling strategy that hierarchically filters redundant temporal information, combined with an adaptive slot channel modeling mechanism that dynamically clusters correlated channels. This design strictly prevents the introduction of spurious correlations among independent variables. Details about the MFE are provided in Section \ref{multi-domain feature extractor}. 

\textbf{Multi-Task Collaborative Training (MCT):} 
To address task diversity, we propose the MCT paradigm, which leverages task specificity and commonality. Instead of pre-training entirely separate models for different tasks, TimeSeek incorporates lightweight, task-customized modules: a global encoder for classification, a non-autoregressive decoder for forecasting, and a patch-masking reconstruction module for anomaly detection. These modules are built on top of the shared MFE representations and are collaboratively optimized via a dual-loss mechanism (prediction loss and reconstruction loss). This symbiotic optimization enables distinct pre-training objectives to mutually reinforce the learned representations. Details about MCT are provided in Section \ref{multi-task collaborative training}.

\subsection{Multi-Domain Feature Extractor}
\label{multi-domain feature extractor} 
To address the domain heterogeneity of time series data in multi-domain pre-training and generation, we propose a multi-domain feature extractor (MFE) that adaptively models complex time series patterns based on the characteristics of each domain. It consists of two components: progressive multi-scale modeling and adaptive slot channel modeling. Specifically, the input time series \( \mathbf{X} \) undergoes downsampling through convolution to obtain time series at different scales: \( \mathbf{X}_{\mathrm{c}} \), \( \mathbf{X}_{\mathrm{m}} \), and \( \mathbf{X}_{\mathrm{f}} \). These are progressively fed into their corresponding scale extractors, from coarse to fine scales, for feature extraction and dynamic fusion. Each scale extractor incorporates temporal modeling and adaptive channel modeling to capture complex temporal dependencies and channel correlations, ultimately generating the representations.

\subsubsection{Progressive Multi-scale Modeling}
\label{progressive multi-scale modeling}
Time series data exhibits significant multi-scale characteristics, including local fluctuations, short-term dynamics, and long-term trends. Multi-scale modeling in pre-training can enhance the understanding of patterns and improve generalization capabilities. 
However, directly fusing information at different scales may lead to feature conflicts or redundancy problems, and also ignore the differences in preferences for different scales in different domains, thereby weakening the model's generalization ability. 
To address this, we propose a progressive multi-scale modeling method. Through progressive fusion, the model progressively extracts and integrates features from coarse to fine scales, building a comprehensive representation that retains both global dependencies and fine-grained patterns. To prevent information conflict during this integration, we further introduce an adaptive gating mechanism to dynamically adjust the fusion weights, filtering redundant features and highlighting key scale information based on domain-specific characteristics.

\begin{algorithm}[tb]
\caption{Progressive Multi-Scale Modeling}
\label{alg:multi_scale}
\textbf{Input}: Raw time series $\mathbf{X}$ \\
\textbf{Output}: Final representation $\mathbf{H}$
\begin{algorithmic}[1]
\STATE \textbf{Stage 1: Multi-Scale Construction}
\STATE Obtain different scales $\mathbf{X}_c, \mathbf{X}_m, \mathbf{X}_f$ via Conv (Eq. 4)
\STATE Generate embeddings $\mathbf{E}_c, \mathbf{E}_m, \mathbf{E}_f$ 
\STATE Initialize $\mathbf{E}^{fuse}_c = \mathbf{E}_c$ \quad \textit{(Start with coarse-grained scale)}
\STATE
\FOR{each scale $s \in \{c, m, f\}$}
    \STATE \textbf{Stage 2: Feature Extraction per Scale}
    \STATE $\mathbf{H}_s = \text{AdaptiveSlotChannelModeling}(\mathbf{E}^{fuse}_s)$ \quad \textit{(Algorithm \ref{alg:channel_modeling})}
    \STATE
    \STATE \textbf{Stage 3: Progressive Fusion (Coarse to Fine)}
    \IF{scale $s \neq f$}
        \STATE $\tilde{\mathbf{F}}_s = \text{Interpolate}(\mathbf{H}_s)$ \quad \textit{(Align with next finer scale)}
        \STATE Compute gating weights $\mathbf{G} = \sigma(\mathbf{W}_g [\tilde{\mathbf{F}}_s; \mathbf{E}_{next}] + \mathbf{b}_g)$ (Eq. 5)
        \STATE $\mathbf{E}^{fuse}_{next} = (\mathbf{G} \odot \tilde{\mathbf{F}}_s) + \mathbf{E}_{next}$ 
    \ENDIF
\ENDFOR
\STATE
\STATE \textbf{return} $\mathbf{H} = \mathbf{H}_f$ \quad \textit{(Output fine-grained representation)}
\end{algorithmic}
\end{algorithm}

\textbf{Multi-scale Construction.} Given a raw time series input $\mathbf{X} \in \mathbb{R}^{C \times T}$, we first derive multi-scale representations using convolution networks (Conv) for downsampling, yielding coarse-grained, medium-grained, and fine-grained sequences $\mathbf{X}_c \in \mathbb{R}^{C \times T/4}, \mathbf{X}_m \in \mathbb{R}^{C \times T/2}$. The original input \( \mathbf{X} \) is treated as the fine-grained sequence, represented as \( \mathbf{X}_f \in \mathbb{R}^{C \times T} \), where $C$ and $T$ represent the number of channels and the input length of the original input, respectively.
\begin{equation}
\begin{aligned}
    \mathbf{X}_\mathrm{c} = \mathrm{Conv}(\mathbf{X}, \mathrm{kernel}=4, \mathrm{stride}=4), \\
    \mathbf{X}_\mathrm{m} = \mathrm{Conv}(\mathbf{X}, \mathrm{kernel}=2, \mathrm{stride}=2)
\end{aligned}
\end{equation}

To preserve local semantic information, a patching strategy partitions each scaled sequence into non-overlapping patches of size $P$. Each patch is then embedded along the patch size dimension to generate corresponding embeddings $\mathbf{E}_c \in \mathbb{R}^{C \times N/4 \times D}, \mathbf{E}_m \in \mathbb{R}^{C \times N/2 \times D}, \mathbf{E}_f  \in \mathbb{R}^{C \times N \times D}$.

\textbf{Progressive Fusion with Adaptive Gating.}
As illustrated in Figure 2, the MFE progressively fuses multi-scale features from coarse to fine granularity. This hierarchical integration, including the coarse-to-medium fusion and the medium-to-fine fusion, allows complementary interactions across scales while alleviating feature conflicts caused by direct fusion. 

Specifically, for the coarse-to-medium fusion, the coarse-grained embedding is first encoded by the coarse-scale extractor, which includes temporal modeling and adaptive channel modeling, to capture temporal dependencies and channel correlations, yielding the coarse-grained features  $\mathbf{F}_c \in \mathbb{R}^{C \times N/4 \times D}$. To align with the medium-grained scale, $\mathbf{F}_c$ undergoes interpolation to double its patch count, yielding $\mathbf{\tilde{F}}_c \in \mathbb{R}^{C \times N/2 \times D}$. Then an adaptive gating, which is composed of linear projections and sigmoid activation, dynamically reweights $\mathbf{\tilde{F}}_c$ to suppress redundant features.
\begin{equation}
    \mathbf{G} = \sigma(\mathbf{W}_g[\mathbf{\tilde{F}}_c; \mathbf{E}_m] + \mathbf{b}_g), \quad \mathbf{\hat{F}}_c = \mathbf{G} \odot \mathbf{\tilde{F}}_c ,
\end{equation}
where $\mathbf{W}_g \in \mathbb{R}^{2D \times D}$ and $\mathbf{b}_g \in \mathbb{R}^{D}$ represent the parameters of linear projections in gating mechanism, $\mathbf{G} \in \mathbb{R}^{C \times N/2 \times D}$ represents the weights of $\mathbf{\tilde{F}}_c $ and $\sigma$ denotes the sigmoid activation. Then the gated coarse features $\mathbf{\hat{F}}_c$ are fused with the medium-grained embeddings $\mathbf{E}_m$ via addition, producing enhanced medium-grained embeddings $\mathbf{E}_m^{\mathrm{fuse}} \in \mathbb{R}^{C \times N/2 \times D}$.

For the medium-to-fine fusion, we adopt the same operation as used in the coarse-to-medium fusion, taking the embeddings $\mathbf{E}_m^{\mathrm{fuse}}$ as input to generate the fine-grained embeddings $\mathbf{E}_f^{\mathrm{fuse}} \in \mathbb{R}^{C \times N \times D}$. Finally, $\mathbf{E}_f^{\mathrm{fuse}}$ is input into the fine-scale feature extractor to obtain the final representation $\mathbf{H}$.

\subsubsection{Adaptive Slot Channel Modeling}
To further enhance the generalization of time series representations across different domains, we consider another key characteristic of time-series data: channel correlations. However, these channel correlations show significant differences across different domains. To address this, we propose an adaptive slot channel modeling method, which primarily includes channel correlation adaptive detection based on a channel router and a channel adaptive clustering process based on the slot attention mechanism.

\begin{figure*}[t]  
    \centerline{\includegraphics[width=1\linewidth]{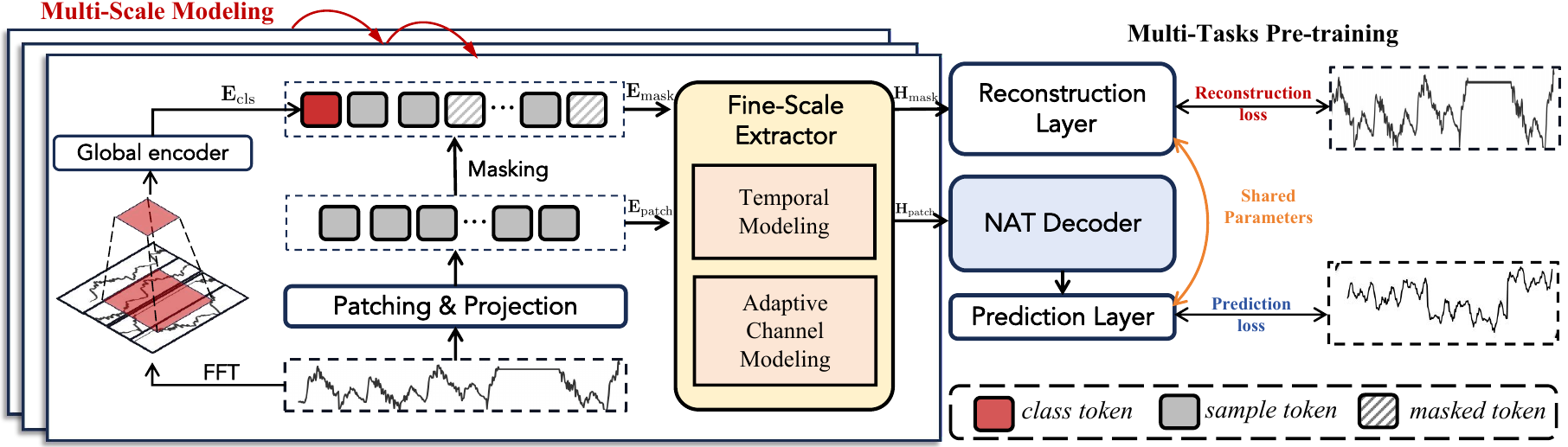}}
    \caption{Multi-Tasks pre-training framework of TimeSeek, where the Fine-Scale Extractor is put on top to ease description.}
    \label{fig: multi-task pretraining}
\end{figure*}

\begin{algorithm}[tb]
\caption{Adaptive Slot Channel Modeling}
\label{alg:channel_modeling}
{\raggedright
\textbf{Input}: Current scale embedding $\mathbf{E}$, Threshold $\tau$, Slots $\mathbf{S}$ \\
\textbf{Output}: Channel-refined representation $\mathbf{H}_c$ \par}
\begin{algorithmic}[1]
\STATE \textbf{Stage 1: Channel Routing}
\STATE Compute dimension-wise average of $\mathbf{E}$ to get simplified embedding $\mathbf{E}_r$
\STATE Calculate routing probability $w$ via softmax
\STATE
\STATE \textbf{Stage 2: Adaptive Clustering \& Masking}
\IF{$w > \tau$}
    \STATE Project $\mathbf{E}$ and $\mathbf{S}$ to obtain $\mathbf{Q}_c$, $\mathbf{K}_c$, $\mathbf{V}_c$ (Eq. 6)
    \STATE Extract slot information $\mathbf{S}' = \text{Softmax}(\mathbf{Q}_c \mathbf{K}_c^T / \sqrt{D})\mathbf{V}_c$ 
    \STATE Update global slots $\mathbf{S}_{new} = \mathbf{S} + \text{MLP}(\mathbf{S} + \mathbf{S}')$ (Eq. 7)
    \STATE Sample slot mask $\mathbf{M}_s$ via Gumbel-Softmax sampling (Eq. 8)
    \STATE Compute masked slots $\mathbf{S}_{mask} = \mathbf{M}_s \odot \mathbf{S}_{new}$
    \STATE $\mathbf{H}_c = \text{Gumbel-Softmax}(\mathbf{E} \cdot \mathbf{S}_{mask}^T / \sqrt{D})\mathbf{S}_{mask}$ (Eq. 9)
\ELSE
    \STATE $\mathbf{H}_c = \text{TemporalModeling}(\mathbf{E})$ \quad \textit{(Bypass channel modeling)}
\ENDIF
\STATE
\STATE \textbf{return} $\mathbf{H}_c$
\end{algorithmic}
\end{algorithm}

\textbf{Channel Router.} To prevent performance degradation caused by unnecessary channel modeling in time series that lack significant channel correlations, we introduce a channel router that dynamically determines whether meaningful channel correlations exist in the input. Specifically, taking the fused medium-grained embedding \( \mathbf{E}_{\mathrm{m}}^{\mathrm{fuse}} \) as an example, the router first generates simplified embedding \( \mathbf{E}_\mathrm{r} \in \mathbb{R}^D \) through dimension-wise average operations: patch-wise and channel-wise averaging. The embedding \( \mathbf{E}_\mathrm{r} \) is then projected through linear layers and a softmax operation to compute the routing probability \( w \), where \( w \) represents the probabilities of activating channel modeling. A predefined threshold \( \tau \) controls the subsequent operations: when \( w > \tau \), the model activates adaptive channel modeling; otherwise, it bypasses this process, and the router directly adds it to the representations obtained by temporal modeling, avoiding redundant computations and potential noise amplification.

\textbf{Channel Slot Attention.} Since time series data often exhibit varying channel correlations across different domains, applying a fixed channel modeling method across all domains may introduce spurious correlations between uncorrelated channels, leading to suboptimal generalization. To enhance cross-domain adaptability, we propose an innovative Channel Slot Attention mechanism. By dynamically clustering channels, strongly correlated channels are grouped within the same cluster while preventing unnecessary modeling of independent channels. This adaptive clustering mechanism not only preserves essential channel interactions but also mitigates domain-specific biases.

For channel correlation modeling, we initialize \( K \) slots \( \mathbf{S} \in \mathbb{R}^{K \times N \times D} \) (\( K < C \), where \( C \) denotes the number of channels and $N$ represents the number of patches.).
These slots not only serve as dynamic memory units to store and update channel interaction patterns but also provide a shared slot space for channels in different domains, thereby reducing dependence on specific domains and better adapting to channel correlations in different domains.
At each patch, the Slot Attention mechanism automatically aggregates cross-channel information through competitive attention weights, allowing for iterative refinement of slot representations. Specifically, we apply linear transformations to both the slots \( \mathbf{S} \) and the input embedding \( \mathbf{E}_{\mathrm{m}}^{\mathrm{fuse}} \) to obtain the Query \( \mathbf{Q}_\mathrm{c} \in \mathbb{R}^{K \times N \times D} \), Key \( \mathbf{K}_\mathrm{c} \), and Value \( \mathbf{V}_\mathrm{c} \in \mathbb{R}^{C \times N \times D} \). Then, the Channel Slot Attention mechanism uses the slots as queries to selectively focus on the input embedding, producing the channel slot information \( \mathbf{S'} \in \mathbb{R}^{K \times N \times D} \).
\begin{equation}
    \begin{aligned}
        \mathbf{Q}_\mathrm{c} = \mathbf{W}_{Q}  \mathbf{S}, \quad
        \mathbf{K}_\mathrm{c} =  & \mathbf{W}_{K}   \mathbf{E}_{\mathrm{m}}^{\mathrm{fuse}}, \quad 
        \mathbf{V}_\mathrm{c} = \mathbf{W}_{V}  \mathbf{E}_{\mathrm{m}}^{\mathrm{fuse}}, \\
        \mathbf{S'} = \mathrm{Softmax} & (\mathbf{Q}_\mathrm{c} \mathbf{K}_\mathrm{c}^T / \sqrt{D}) \mathbf{V}_\mathrm{c}, \\
    \end{aligned}
\end{equation}
where $\mathbf{W}_Q$, $\mathbf{W}_K$ and $\mathbf{W}_V \in \mathbb{R}^{D \times D}$ represent learnable projections matrices. Next, we combine the obtained channel slot information \( \mathbf{S'} \) with the previous slots \( \mathbf{S} \) using a Multi-Layer Perceptron (MLP). This combination is then added to the original slots \( \mathbf{S} \) to update the slots, resulting in the updated slots \( \mathbf{S}_{\mathrm{new}} \in \mathbb{R}^{K \times N \times D} \).
\begin{equation}
    \mathbf{S}_{\mathrm{new}} = \mathbf{S} + \mathrm{MLP}(\mathbf{S}+\mathbf{S'})
\end{equation}

To avoid all slots focusing on similar patterns or redundant information, we introduce a masking strategy to dynamically filter less informative slots and retain important ones, while adaptively selecting the number of slots. We employ a lightweight MLP based on the $\mathbf{S}_{\mathrm{new}}$ to generate a probability matrix $\Pi \in \mathbb{R}^{K \times 1}$ for slots, indicating whether each slot should be retained or masked. Here, $\Pi_{i,0}$ represents the probability that the $i$-th slot is retained. We then perform Gumbel-Softmax sampling on the probability matrix $\Pi$ to obtain the slot mask $\mathbf{M}_s$. Based on the Slot mask \( \mathbf{M}_s \), we apply a masking operation to \( \mathbf{S}_{\mathrm{new}} \), resulting in \( \mathbf{S}_{\mathrm{mask}} \), where \( \mathbf{S}_{\mathrm{mask}} \) denotes the key channel cluster information.
\begin{equation}
    \mathbf{M}_{\mathrm{s}} = \mathrm{GumbelSoftmax(\mathbf{\Pi)}}, \quad \mathbf{S}_{\mathrm{mask}} = \mathbf{M}_{\mathrm{s}} \odot \mathbf{S}_{\mathrm{new}}
\end{equation}

Subsequently, we compute attention weights between the masked slots $\mathbf{S}_{\mathrm{mask}}$ and input embeddings $\mathbf{E}_{\mathrm{m}}^{\mathrm{fuse}}$, considering that directly using weighted summation may cause the channel representation to mix the information of multiple slots, making it blurred. We use Gumbel-Softmax sampling and $\mathbf{S}_{\mathrm{mask}}$ to produce the final channel representation $\mathbf{H}_{c} \in \mathbb{R}^{C \times N \times D}$.
\begin{equation}
    \mathbf{H}_{\mathrm{c}} = \mathrm{GumbelSoftmax(\mathbf{E}_{\mathrm{m}}^{\mathrm{fuse}} \cdot \mathbf{S}_{\mathrm{mask}}^T / \sqrt{D}) \mathbf{S}_{\mathrm{mask}}}
\end{equation}

\subsection{Multi-Task Collaborative Training}
\label{multi-task collaborative training}

Considering that time series tasks are diverse in downstream scenarios, designing a foundation model that can support multi-task learning becomes essential. However, time series tasks exhibit both heterogeneity and commonality: task heterogeneity may lead to negative interference between tasks, while task commonality provides potential synergies for collaborative learning. To address these, we propose multi-task collaborative training, including task customization and cross-task training.  In task customization, we design task-specific modules based on the unique characteristics of each task, enhancing the capabilities in prediction generation, anomaly detection, and categorical discrimination. For cross-task training, we further propose unified components for shared feature extraction and loss computation, strengthening the ability to learn task commonality.

\vspace{0em}
\subsubsection{Task Customization}

We design task-specific customization modules for classification, forecasting, and anomaly detection to better capture the unique characteristics of each task while mitigating negative interference. This design improves the capacity to handle diverse tasks in a unified framework.

As illustrated in Figure \ref{fig: multi-task pretraining}, multi-scale modeling refers to Section \ref{progressive multi-scale modeling}, which is stacked together. 
Take the Fine-scale Extractor as an example, 
we perform non-overlapping patch segmentation to obtain $\mathbf{X}_{\mathrm{patch}} \in \mathbb{R}^{C \times N \times P}$, followed by a linear projection to produce the embeddings $\mathbf{E}_{\mathrm{patch}} \in \mathbb{R}^{C \times N \times D}$.

\textbf{Customization for Classification.} 
Motivated by the inherent reliance of classification on global context to distinguish categories, we propose a novel module designed to extract these global representations and enhance the model's discriminative capability.

The input time series $\mathbf{X}$ is first transformed into the frequency domain using the Fast Fourier Transform.  The dominant frequency $f$ is identified by selecting the component with the highest amplitude, from which we compute the corresponding period $P_d$.
\begin{equation}
A = \mathrm{Avg}(\mathrm{Amp}(\mathrm{FFT}(\mathbf{X}))), \quad f = \mathrm{arg} \mathrm{Top1}(A), \quad P_d = \lceil \frac{T}{f} \rceil
\end{equation}
Based on the period $P_d$, the time series is transformed from a 1D temporal space into a 2D temporal space, resulting in $\mathbf{X}_\mathrm{d} \in \mathbb{R}^{C \times f \times P_d}$. To better highlight the temporal differences for classification, we apply a linear resizing operation along the period and frequency dimensions, yielding $\mathbf{X}_\mathrm{r} \in \mathbb{R}^{C \times D_r \times D_r}$, where $D_r$ represents the feature dimension after resize. The resized time series $\mathbf{X}_\mathrm{r}$ is passed through the global encoder to extract discriminative features, producing the classification embedding $\mathbf{E}_\mathrm{c}$. Finally, this embedding is flattened and projected to generate the class token $\mathbf{E}_\mathrm{cls} \in \mathbb{R}^{C \times D}$, which is fed into the feature extractor to obtain the final classification representation $\mathbf{H}_\mathrm{cls} \in \mathbb{R}^{C \times D}$.

\textbf{Customization for Forecasting.} To enhance the generative ability for forecasting, we propose a Non-Autoregressive Forecasting module, including a Non-Autoregressive Transformer Decoder (NAT Decoder) and Prediction Layer, that generates all predictions at once, avoiding the error accumulation inherent in autoregressive methods. 

After patching, the time series is progressively fed into the corresponding Scale Feature Extractor to model complex patterns at different scales, yielding representations \( \mathbf{H}_{\mathrm{patch}} \in \mathbb{R}^{C \times N \times D} \).
Considering that the final patch of the representation retains the overall information of the input sequence and is closest to the future prediction sequence, we replicate the last patch of the representations $\mathbf{H}_{\mathrm{patch}}$ $M$ times as the input and inject Rotary Position Embedding (RoPE) to the NAT Decoder, denoted as $\mathbf{H}_{\mathrm{d}} \in \mathbb{R}^{C \times M \times D}$, where $M$ corresponds to the prediction patch count during pre-training. Subsequently, $\mathbf{H}_{\mathrm{d}}$ is passed through causal attention, followed by cross-attention with $\mathbf{X}_{\mathrm{patch}}$, and is refined via a feedforward layer to yield prediction representation $\mathbf{H}_\mathrm{f} \in \mathbb{R}^{C \times M \times D}$. Finally, the prediction layer is applied, followed by flattening, to produce the predicted sequence $Y_{f} \in \mathbb{R}^{C \times F}$.

\textbf{Customization for Anomaly Detection.} 
To enhance the model's capability to capture localized anomalies, we introduce a patch-masking reconstruction module.  

Given the patch embedding $\mathbf{E}_{\mathrm{patch}}$, we apply random masking along the patch dimension to produce a masked embedding $\mathbf{E}_{\mathrm{mask}} \in \mathbb{R}^{C \times N \times D}$.
Considering that anomaly detection needs global information, we concatenate the masked embedding \(\mathbf{E}_{\mathrm{mask}}\) with the class token obtained from the global encoder along the patch dimension to form the new masked embedding \(\mathbf{E'}_{\mathrm{mask}} \in \mathbb{R}^{C \times (N+1) \times D}\). The embedding is then progressively input into its corresponding scale extractor to model time series patterns, yielding the representation \(\mathbf{H}_{\mathrm{mask}}\), which is finally decoded by the reconstruction layer to get the reconstructed output $\mathbf{Y}_{\mathrm{r}} \in \mathbb{R}^{C \times N \times P}$.

\subsubsection{Cross-task Training}
Our innovative architectural design effectively addresses the requirements of multi-task pre-training by mitigating negative interference across diverse objectives while enhancing predictive generation, anomaly detection, and categorical discrimination capabilities. Considering both the unique requirements and inherent commonalities among different tasks, we further propose unified components for shared feature extraction and loss computation to facilitate collaborative cross-task learning. 
This approach enables TimeSeek to leverage complementary information while preserving task-specific advantages.

At the structural level, all tasks share a common multi-domain feature extractor that comprehensively models time series patterns, learning generalized representations through mutual collaboration across tasks. For classification and anomaly detection, we concatenate the class token from the global encoder with masked input sequences. This combination injects global semantic information into the reconstruction process, providing valuable prior knowledge for anomaly identification.
Simultaneously, the reconstruction pre-training not only reciprocally updates the classification module but also deepens the model’s structured understanding of time series, enabling the downstream classification task to leverage richer representations for more accurate discrimination.

At the loss level, TimeSeek employs a dual-loss mechanism comprising prediction loss $\mathcal{L}_{\mathrm{pred}}$ and reconstruction loss $\mathcal{L}_{\mathrm{recon}}$ to optimize model parameters. The prediction loss strengthens the model's generation capability through non-autoregressive forecasting objectives, particularly beneficial for few-shot and zero-shot prediction scenarios.
The reconstruction loss focuses on anomaly detection by enforcing precise recovery of masked tokens, while its gradient signals directly enhance the classification module's feature extraction. This symbiotic optimization ensures the learned representations maintain high adaptability for anomaly detection and classification tasks. In contrast, directly applying the prediction loss to optimize classification could lead to over-smoothing of critical discriminative features, ultimately resulting in conflicting representations between the tasks.
\begin{equation}
   \mathcal{L}_{\mathrm{pre-train}} =\mathcal{L}_{\mathrm{pred}} +\mathcal{L}_{\mathrm{recon}}
\end{equation}
Through these designed interactions, TimeSeek makes different pre-training objectives mutually reinforce each other, ultimately producing robust and generalized representations for diverse tasks.

\textbf{In the downstream scenarios}, we fine-tune the pre-trained model or perform direct inference for each task. For classification, the time series is encoded by the Global Encoder to extract a class token, which is then processed by the MFE to obtain $\mathbf{H}_{\mathrm{cls}}$. These representations are fed into a downstream classifier to produce the final classification results. For forecasting, the time series is processed by the MFE to obtain representations $\mathbf{H}_{\mathrm{patch}}$, which are fed into the NAT decoder and prediction layer to generate the output $\mathbf{Y}_{\mathrm{f}}$. The model is optimized using prediction loss.  For anomaly detection, the time series undergoes patching and masking before being processed by the MFE to get $\mathbf{H}_{\mathrm{mask}}$. These representations are fed to the reconstruction layer to produce $\mathbf{Y}_{r}$, with the model updated by reconstruction loss.

\section{Experiment}
\subsection{Experimental Setup}
\subsubsection{Datasets}
We perform pre-training and evaluation of TimeSeek on diverse datasets and \textit{ensure no overlap between the pre-training datasets and the evaluation datasets}. More specific information about datasets is in the Supplemental Material.

\textbf{Pre-train Datasets.}
To achieve general time series analysis, we compile an extensive array of publicly accessible datasets for pre-training, encompassing various domains such as energy, nature, transportation, web, health, and economics. In total, the pre-training data comprises approximately 1 billion time series data points.

\textbf{Evaluation Datasets.} To validate TimeSeek under multi-task, we use the widely used evaluation datasets for forecasting, anomaly detection, and classification tasks, which also span multiple domains to validate the generalizability and versatility of TimeSeek. In our evaluation, we adopt the same train-validation-test split as the baselines for each task to ensure a fair comparison.

\begin{itemize}
\item \textbf{Forecasting.} We conduct experiments for forecasting on eight widely recognized forecasting benchmarks, including ETT~(ETTm1, ETTm2, ETTh1, ETTh2), Weather, Traffic, Solar, Exchange, and GIFT-Eval Benchmark.

\item \textbf{Anomaly Detection.} We evaluate five widely used anomaly detection benchmarks: SMD, MSL, SMAP, SWaT, and PSM.

\item \textbf{Classification.} We follow existing works~\cite{Timesnet} to use 10 datasets for UEA, which feature from 2 to 26 categories for the experiments of classification,
including EthanolConcentration, FaceDetection, Handwriting, Heartbeat, JapaneseVowels, PEMS-SF, SelfRegulationSCP1, SelfRegulationSCP2, SpokenArabicDigits, and UWaveGestureLibrary.
\end{itemize}

\subsubsection{Evaluation Setting}  

We partition the downstream evaluation dataset into training, validation, and test sets. The evaluation settings consist of the following three scenarios: \textbf{Zero-shot Setting}: The model is directly applied to the test set without fine-tuning, assessing its generalization capability of the foundation models.
\textbf{Few-shot Setting}: A small subset of the training data, along with the entire validation set, is used for fine-tuning, evaluating the model's performance in few-shot scenarios. \textbf{Full-shot Setting}: The model is trained using the entire training and validation datasets, measuring its performance under ample data conditions. \textit{In our experiments, TimeSeek is evaluated under the zero-shot and few-shot settings, while the comparison time-series-specific model is trained and tested in the full-shot setting.
}

\subsubsection{Baselines} We select both time series foundation models and time-series-specific models of each task as baselines.

\textbf{Time Series Foundation Baselines.} 
Since existing time series foundation models primarily focus on time series forecasting tasks, we select six state-of-the-art foundation models for comparison in a zero-shot forecasting setting, including Timer~\cite{timer}, MOIRAI~\cite{moirai}, Chronos~\cite{chronos},
TimesFM~\cite{timesfm}, UniTS~\cite{units}, Time-MOE~\cite{timemoe},TEMPO~\cite{tempo}. 

\textbf{Forecasting Baselines.}
We select six SOTA time series forecasting models as baselines, including iTransformer~\cite{iTransformer}, FITS~\cite{fits}, TimeMixer~\cite{timemixer}, PatchTST~\cite{patchtst}, Crossformer~\cite{fits}, and DLinear~\cite{dlinear}.

\textbf{Anomaly Detection Baselines.} We compare TimeSeek with 12 anomaly detection baselines for comprehensive evaluations, including the linear transformation-based model OCSVM; the outlier-based methods IForest; the neural network-based models: AutoEncoder (AE)~\cite{ae1}, DAGMM~\cite{DAGMM}, LSTM~\cite{LSTM-NDT}, BeatGAN~\cite{gan1}, OmniAnomaly (Omni)~\cite{OmmiAnomaly}, Anomaly Transformer (A.T.)~\cite{A-T}, DCdetector~\cite{DCdetector}, D3R~\cite{D3R}, and ModernTCN~\cite{moderntcn}.

\textbf{Classification Baselines.} 
We compare TimeSeek with nine baselines, including traditional classification methods: DTW~\cite{DTW}, XGBoost~\cite{XGBoost}, and Rocket~\cite{Rocket}, as well as deep learning-based methods: OS-CNN~\cite{OS-CNN}, TapNet~\cite{tapnet}, PatchTST~\cite{patchtst}, TimesNet~\cite{Timesnet}, DLinear~\cite{dlinear}, and Crossformer~\cite{crossformer}, Moment~\cite{moment}.

\subsubsection{Metrics} Following prior studies, we use Mean Squared Error (MSE) and Mean Absolute Error (MAE) as evaluation metrics for the forecasting task. The lower the metric values, the better the prediction accuracy. We use Affiliation-based F1-score (F1), a widely used metric for anomaly detection that balances affiliated precision (P) and recall (R) \cite{local_evaluation}. The higher the metric values, the better the anomaly detection accuracy. 
Average Accuracy (Avg. ACC) and the count of datasets achieving the highest accuracy (Num. Top-1) as evaluation metrics for classification. The higher the metric values, the better the classification accuracy.

\subsubsection{Experiment Settings and Implementations}

During pre-training, TimeSeek employs the Adam optimizer with an initial learning rate of 0.0005 and undergoes 30 epochs of training, incorporating an early stopping strategy with a patience of 10 epochs. The batch size is set to 2048, and the input time series length is 576.  For the forecasting task, all methods predict future values with lengths \( F=\{96, 192, 336, 720\} \) to ensure a fair comparison. Additionally, all models do not employ the drop-last strategy \cite{tfb}.  For the anomaly detection task, we use a sliding window approach with a window size of 96. After obtaining the anomaly scores, we apply the widely used SPOT method, following existing works \cite{D3R}, to determine the threshold for classifying each point as an outlier. All experiments are implemented with PyTorch and the pre-training experiment is executed on four NVIDIA Tesla-A800-80GB GPUs.

\begin{table*}[htbp]
  \centering
  \caption{The zero-shot forecasting results of time series foundation models. The average results of all predicted lengths $\{96, 192, 336, 720\}$ are listed here. A dash (’-’) denotes datasets included in the model’s pretraining and therefore excluded from testing. ETTm represents the average results across ETTm1 and ETTm2, while ETTh represents the average across ETTh1 and ETTh2. Bold: the best, and underline: the second best.}
  \resizebox{\textwidth}{!}{
    \setlength{\tabcolsep}{3pt} 
    \begin{tabular}{l cc cc cc cc cc cc cc cc}
      \hline 
      \hline 
      Method & \multicolumn{2}{c}{TimeSeek} & \multicolumn{2}{c}{Timer} & \multicolumn{2}{c}{TimesFM} & \multicolumn{2}{c}{Chronos} & \multicolumn{2}{c}{Time-MoE} & \multicolumn{2}{c}{UniTS} & \multicolumn{2}{c}{MOIRAI} & \multicolumn{2}{c}{TEMPO} \\
      \cmidrule(lr){2-3} \cmidrule(lr){4-5} \cmidrule(lr){6-7} \cmidrule(lr){8-9} \cmidrule(lr){10-11} \cmidrule(lr){12-13} \cmidrule(lr){14-15} \cmidrule(lr){16-17}
      Metric & MSE & MAE & MSE & MAE & MSE & MAE & MSE & MAE & MSE & MAE & MSE & MAE & MSE & MAE & MSE & MAE \\
      \midrule
      ETTm(Avg) & \textbf{0.311} & \textbf{0.350} & 0.636 & 0.515 & 0.391 & 0.389 & 0.422 & 0.392 & \underline{0.322} & 0.368 & 0.548 & 0.457 & 0.333 & \underline{0.354} & 0.391 & 0.393 \\
      ETTh(Avg) & \textbf{0.390} & \textbf{0.406} & 0.466 & 0.441 & 0.439 & 0.422 & 0.462 & 0.424 & \underline{0.391} & \underline{0.413} & 0.394 & 0.416 & 0.432 & 0.423 & 0.394 & 0.413 \\
      Weather   & \textbf{0.217} & \textbf{0.250} & 0.292 & 0.313 & --    & --    & 0.288 & 0.309 & 0.256 & 0.289 & 0.275 & 0.298 & \underline{0.260} & \underline{0.275} & 0.282 & 0.316 \\
      Traffic   & \underline{0.555} & \underline{0.367} & 0.613 & 0.407 & --    & --    & 0.615 & 0.421 & --    & --    & --    & --    & --    & --    & \textbf{0.503} & \textbf{0.358} \\
      Solar     & \textbf{0.192} & \textbf{0.257} & 0.771 & 0.604 & 0.500 & 0.397 & \underline{0.393} & \underline{0.319} & 0.411 & 0.428 & 0.845 & 0.669 & 0.714 & 0.704 &     0.416    & 0.374    \\
      Exchange  & \textbf{0.341} & \textbf{0.390} & 0.392 & 0.425 & 0.386 & 0.417 & 0.370 & 0.412 & 0.390 & 0.417 & 0.432 & 0.454 & \underline{0.354} & \underline{0.400} & 0.398    & 0.426    \\
      \hline
      \hline 
    \end{tabular}
  }
  \label{tab:forecasting_zero_shot}%
\end{table*}
\begin{table*}[htbp]
  \centering
  \caption{The forecasting results of TimeSeek and time series-specific models are presented, with average performance reported across all prediction lengths $\{96, 192, 336, 720\}$. TimeSeek utilizes only 10\% of the training data for few-shot learning, whereas time series-specific models are trained using the full training data. Bold: the best and underline: the second best. Full results are in the Supplemental Material.
  }
  \resizebox{\textwidth}{!}{
  \setlength{\tabcolsep}{2.0mm}{
    \begin{tabular}{cc|cc|cc|cc|cc|cc|cc|cc}
    \hline
    \hline
     \multicolumn{2}{c|}{Method} & \multicolumn{2}{c|}{TimeSeek \textcolor{red}{(10\%)}} & \multicolumn{2}{c|}{iTransformer} & \multicolumn{2}{c|}{FITS} & \multicolumn{2}{c|}{TimeMixer} & \multicolumn{2}{c|}{PatchTST} & \multicolumn{2}{c|}{Crossformer} & \multicolumn{2}{c}{DLinear} \\
    \midrule
    \multicolumn{2}{c|}{Metric} & \multicolumn{1}{c}{MSE} & \multicolumn{1}{c|}{MAE} & \multicolumn{1}{c}{MSE} & \multicolumn{1}{c|}{MAE} & \multicolumn{1}{c}{MSE} & \multicolumn{1}{c|}{MAE} & \multicolumn{1}{c}{MSE} & \multicolumn{1}{c|}{MAE} & \multicolumn{1}{c}{MSE} & \multicolumn{1}{c|}{MAE} & \multicolumn{1}{c}{MSE} & \multicolumn{1}{c|}{MAE} & \multicolumn{1}{c}{MSE} & \multicolumn{1}{c}{MAE} \\
    \midrule
    \multicolumn{2}{c|}{ETTm(avg)} & \textbf{0.285} & \textbf{0.337} & \underline{0.303} & 0.348 & 0.306 & \underline{0.345} & 0.307 & 0.349 & \underline{0.303} & 0.348 & 0.481 & 0.479 & 0.307 & 0.350 \\
    \multicolumn{2}{c|}{ETTh(avg)} & \textbf{0.365} & \textbf{0.394} & 0.400 & 0.421 & \underline{0.372} & \underline{0.407} & 0.387 & 0.418 & 0.385 & 0.416 & 0.666 & 0.57  & 0.456 & 0.455 \\
    \multicolumn{2}{c|}{Weather} & \textbf{0.210} & \textbf{0.246} & 0.232 & 0.270 & 0.244 & 0.281 & 0.225 & 0.263 & \underline{0.224} & \underline{0.261} & 0.233 & 0.292 & 0.239 & 0.290 \\
    \multicolumn{2}{c|}{Traffic} & 0.413 & \textbf{0.273} & \textbf{0.397} & 0.281 & 0.429 & 0.302 & 0.410 & 0.279 & \textbf{0.397} & \underline{0.275} & 0.521 & 0.282 & 0.434 & 0.295 \\
    \multicolumn{2}{c|}{Solar} & \textbf{0.188} & \textbf{0.253} & 0.202 & \underline{0.260} & 0.232 & 0.268 & 0.203 & 0.261 & \underline{0.200} & 0.284 & 0.204 & 0.232 & 0.247 & 0.309 \\
    \multicolumn{2}{c|}{Exchange} & 0.332 & \textbf{0.384} & \textbf{0.321} & \textbf{0.384} & 0.349 & 0.396 & 0.385 & 0.418 & \underline{0.322} & \underline{0.385} & 0.389 & 0.423 & 0.349 & 0.411 \\
    \hline
    \hline
    \end{tabular}%
    }
    }
  \label{tab:few-shot forecasting}%
\end{table*}%

\begin{table*}[hbtp]
    \caption{Performance of TimeSeek on the GIFT-Eval leaderboard. A lower MASE or CRPS indicates a better performance. Baseline results are from the GIFT-Eval leaderboard.}
    \vspace{-1em}
    \label{tab:gift_eval}
    \centering
    \renewcommand{\multirowsetup}{\centering}
    \setlength{\tabcolsep}{5.1pt}
    \resizebox{\linewidth}{!}{
    \begin{tabular}{l|cccccccccccccc}
    \hline 
    \hline 
    \textbf{Type} & \multicolumn{4}{c}{\textbf{Statistical Methods}} & \multicolumn{5}{c}{\textbf{Task-Specific Models (Superwised)}} & \multicolumn{5}{c}{\textbf{Time Series Foundation Models (Zero-Shot)}} \\
    \cmidrule(lr){1-1} \cmidrule(lr){2-5} \cmidrule(lr){6-10} \cmidrule(lr){11-15}
        \multirow{2}{*}{\textbf{Model}}
            & \multirow{2}{*}{{Na\"ive}} & {Auto} & {Auto} & {Auto} & {DeepAR} & {TiDE} & {N-BEATS} & {PTST.} & {DLinear} & {TimesFM} & {TabPFN} & {Chronos} & {MOIRAI} & \textbf{TimeSeek} \\
            & & {ETS} & {ARIMA} & {Theta} & (2020) & (2023) & (2019) & (2022) & (2023) & (2023) & (2025) & (2024) & (2024) &  \textbf{(Ours)} \\
        \midrule
            \textbf{MASE} & 1.270 & 1.212 & 1.074 & 1.090 & 1.343  & 1.091 & 0.938 & 0.849 & 1.061 & 1.077 & \underline{0.771} & 0.822 & 0.875 &  \textbf{0.732}\\
            \textbf{CRPS}  & 1.591 & 7.489 & 0.912 & 1.244 & 0.853 & 0.772 & 0.816 & 0.587 & 0.846 & 0.680 & \underline{0.544} & 0.577 & 0.599 &  \textbf{0.528}\\
    \hline 
    \hline 
    \end{tabular}}
\end{table*}

\subsection{Main Results}
\subsubsection{Forecasting Results} 
 For a fair comparison, we perform zero-shot predictions for each foundational model on downstream datasets that are not part of its pre-training datasets. As shown in Table~\ref{tab:forecasting_zero_shot},
 TimeSeek shows leading performance on all datasets, surpassing the previous best method, Time-MOE, by a clear margin and achieving an average MSE reduction of 16.9\% and MAE reduction of 13.3\%, demonstrating the generalization of the model under different downstream datasets. 
Compared to forecasting foundation models such as Timer, TimesFM, and Chronos, TimeSeek achieves superior forecasting accuracy. This highlights the effectiveness of its multi-task pre-training, which enhances the model’s generalization ability in forecasting tasks. Additionally, compared to UniTS, another multi-task learning model, TimeSeek achieves 32.7\%  and 23.5\%  reduction on the MSE and MAE metrics, further underscoring the importance of considering both task commonality and task specificity in multi-task learning.  Moreover, TimeSeek also outperforms the channel modeling approach MOIRAI, demonstrating that its adaptive channel modeling for multi-domain data effectively improves generalization.

Table \ref{tab:few-shot forecasting} presents the performance of TimeSeek in the 10\% few-shot setting compared to time-series-specific forecasting baselines in the full-shot setting. Despite using significantly less train data from downstream datasets, TimeSeek achieves superior prediction accuracy over these strong baselines. This result highlights TimeSeek’s effectiveness in data-limited scenarios and demonstrates its strong generalization and transfer ability. Compared to single-dataset multi-scale methods like TimeMixer and Crossformer, TimeSeek effectively adapts to multi-domain data through progressive multi-scale modeling, enabling more accurate predictions.

\subsubsection{Anomaly Detection Results}
To ensure a fair comparison in the anomaly detection task, we apply a consistent thresholding protocol across all methods. Unlike baseline models that are trained on the full dataset for each specific scenario, TimeSeek performs zero-shot anomaly detection without any fine-tuning on downstream data, directly conducting inference.
As shown in Table \ref{anomaly detection results}, TimeSeek achieves state-of-the-art results on most datasets without requiring fine-tuning on downstream data. This demonstrates that TimeSeek has a robust general detection capability through extensive multi-domain pre-training, effectively distinguishing between various normal and anomalous patterns.

\subsubsection{Classification Results}
Table \ref{tab:classfication_few_shot} presents the classification results of TimeSeek in the 10\% few-shot setting compared to time-series-specific classification models in the full-shot setting. Remarkably, despite utilizing significantly less training data from downstream datasets, TimeSeek achieves top performance on 6 out of 10 datasets, with an overall accuracy of 0.794. This represents an average accuracy improvement of 7.3\% over the second-best model, TimesNet. These results highlight TimeSeek's exceptional adaptability and generalization capability across diverse classification scenarios, even with limited data.
\begin{table*}
    \centering
    \caption{The anomaly detection results are presented. TimeSeek performs zero-shot inference without utilizing any training data, whereas baselines are trained on the full training data.
    Bold: the best and underline: the second best. More metric results are in the Supplemental Material.}
    \resizebox{\linewidth}{!}{
\begin{tabular}{c|ccc|ccc|ccc|ccc|ccc}
\hline \hline
Dataset
& \multicolumn{3}{c|}{SMD} 
& \multicolumn{3}{c|}{MSL} 
& \multicolumn{3}{c|}{SMAP}
& \multicolumn{3}{c|}{SWaT}
& \multicolumn{3}{c}{PSM} \\ 
\midrule
Metric
& P & R & F1
& P & R & F1
& P & R & F1
& P & R & F1 
& P & R & F1 \\ 
\midrule
OCSVM 
& 66.98 & 82.03 & 73.75 & 50.26 & 99.86 & 66.87 & 41.05 & 69.37 & 51.58 & 56.80 & 98.72 & 72.11  & 57.51 & 58.11 & 57.81  
\\
IForest  
& 71.94 & 94.27 & 81.61 & 53.87 & 94.58 & 68.65 & 41.12 & 68.91 & 51.51 & 53.03 & 99.95 & 69.30 & 69.66 & 88.79 & 78.07  
\\
AE  
& 69.22 & 98.48 & 81.30 & 55.75 & 96.66 & 70.72 & 39.42 & 70.31 & 50.52 & 54.92 & 98.20 & 70.45 & 60.67 & 98.24 & 75.01
\\
DAGMM  
& 63.57 & 70.83 & 67.00 & 54.07 & 92.11 & 68.14  & 50.75 & 96.38 & 66.49  & 59.42 & 92.36 & 72.32 & 68.22 & 70.50 & 69.34 
\\
LSTM   
& 60.12 & 84.77 & 70.35 & 58.82 & 14.68 & 23.49 & 55.25 & 27.70 & 36.90 & 49.99 & 82.11 & 62.15 & 57.06 & 95.92 & 71.55  
\\
BeatGAN   
& 74.11 & 81.64 & 77.69 & 55.74 & 98.94 & 71.30 & 54.04 & 98.30 & 69.74 & 61.89 & 83.46 & 71.08 & 58.81 & 99.08 & 73.81  
\\
Omni   
& 79.09 & 75.77 & 77.40 & 51.23 & 99.40 & 67.61 & 52.74 & 98.51 & 68.70 & 62.76 & 82.82 & 71.41 & 69.20 & 80.79 & 74.55 
\\
CAE-Ensemble  
& 73.05 & 83.61 & 77.97 & 54.99 & 93.93 & 69.37 & 62.32 & 64.72 & 63.50 & 62.10 & 82.90 & 71.01 & 73.17 & 73.66 & 73.42 
\\
A.T. 
& 54.08 & 97.07 & 66.42 & 51.04 & 95.36 & 66.49 & 56.91 & 96.69 & 71.65 & 53.63 & 98.27 & 69.39 & 54.26 & 82.18 & 65.37 
\\
DCdetector  
& 50.93 & 95.57 & 66.45 & 55.94 & 95.53 & 70.56 & 53.12 & 98.37 & 68.99 & 53.25 & 98.12 & 69.03  & 54.72 & 86.36 & 66.99  
\\
D3R  
& 64.87 & 97.93 & 78.02 & 66.85 & 90.83 & 77.02 & 61.76 & 92.55 & \underline{74.09} & 60.14 & 97.57 & \textbf{74.39} & 73.32 & 88.71 
& \underline{80.29}  
\\
ModernTCN   
& 74.07 & 94.79 & \underline{83.16} & 65.94 & 93.00 & \underline{77.17} & 69.50 & 65.45 & 67.41 & 59.14 & 89.22 & 71.13 & 73.47 & 86.83 & 79.59 
\\
\midrule
TimeSeek (\textcolor{red}{zero shot}) 
& 77.09 & 94.00 & \textbf{84.71} & 68.22 & 94.48 & \textbf{79.24} & 64.98 & 88.06 & \textbf{74.78}   & 58.92  & 94.37  &  \underline{72.55} &  74.58 & 86.55  &  \textbf{80.66}
\\ \hline 
\hline
\end{tabular}
}
\label{anomaly detection results}
\end{table*}

\begin{table*}[htbp]
  \centering
  \caption{The classification results are presented. TimeSeek utilizes only 10\% of the training data for few-shot learning, whereas baselines are trained using the full training data. Bold: the best and underline: the second best.}
      \resizebox{\linewidth}{!}{
    \begin{tabular}{cccccccccccc}
    \hline
    \hline
    \multicolumn{1}{c|}{Dataset} & TimeSeek \textcolor{red}{(10\%)}  & Moment & TimesNet & PatchTST & DLinear & Crossformer & OS-CNN & TapNet & Rocket & XGBoost & \multicolumn{1}{c}{DTW } \\
    \midrule
\multicolumn{1}{c|}{EthanolConcentration} & \textbf{0.718} & 0.300 &  0.357  & 0.328  & 0.362  & 0.380  & 0.240  & 0.323  & 0.447  & 0.437  & \multicolumn{1}{c}{0.323 } \\
\multicolumn{1}{c|}{FaceDetection} & 0.651 & 0.689 & 0.686  & 0.683  & 0.680  & 0.687  & 0.575  & 0.556  & \textbf{0.694} & 0.633  & \multicolumn{1}{c}{0.529} \\
\multicolumn{1}{c|}{Handwriting} & 0.361 & 0.351 & 0.321  & 0.296  & 0.270  & 0.288  & \textbf{0.668} & 0.357  & 0.567  & 0.158  & \multicolumn{1}{c}{0.286} \\
\multicolumn{1}{c|}{Heartbeat} & \textbf{0.814} & 0.737 & 0.780  & 0.749  & 0.751  & 0.776  & 0.489  & 0.751  & 0.718  & 0.732  & \multicolumn{1}{c}{0.717} \\
\multicolumn{1}{c|}{JapaneseVowels} & 0.986  & 0.957 & 0.984  & 0.975  & 0.962  & \textbf{0.991} & 0.991  & 0.965  & 0.965  & 0.865  & \multicolumn{1}{c}{0.949} \\
\multicolumn{1}{c|}{PEMS-SF} & \textbf{1.000} & 0.855 & 0.896  & 0.893  & 0.751  & 0.859  & 0.760  & 0.751  & 0.856  & 0.983  & \multicolumn{1}{c}{0.711} \\
\multicolumn{1}{c|}{SelfRegulationSCP1} & \textbf{0.918} & 0.887 &  0.918  & 0.907  & 0.873  & 0.921  & 0.835  & 0.652  & 0.866  & 0.846  & \multicolumn{1}{c}{0.777 } \\
\multicolumn{1}{c|}{SelfRegulationSCP2} & \textbf{0.594} & 0.550 & 0.572  & 0.578  & 0.505  & 0.583  & 0.532  & 0.550  & 0.514  & 0.489  & \multicolumn{1}{c}{0.539 } \\
\multicolumn{1}{c|}{SpokenArabicDigits} & 0.936 & 0.994 & 0.990  & 0.983  & 0.814  & 0.979  & \textbf{0.997} & 0.983  & 0.630  & 0.696  & \multicolumn{1}{c}{0.963} \\
\multicolumn{1}{c|}{UWaveGestureLibrary} & \textbf{0.959} & 0.900 & 0.853  & 0.858  & 0.821  & 0.853  & 0.927  & 0.894  & 0.944  & 0.759  & \multicolumn{1}{c}{0.903 } \\
\midrule
\multicolumn{1}{c|}{Avg. ACC} &  \textbf{0.794}  & 0.722   & \underline{0.736}  & 0.725  & 0.679  & 0.732  & 0.701  & 0.678  & 0.720  & 0.660  & \multicolumn{1}{c}{0.670} \\
\multicolumn{1}{c|}{Num. Top-1} &   \textbf{6}  & 0  & 0     & 0     & 0     & 1     & \underline{2}     & 0     & 1     &   0   &  0 \\
    \hline
    \hline
    \end{tabular}%
    }
  \label{tab:classfication_few_shot}%
\end{table*}%

\textbf{Summary:} As illustrated in Tables \ref{tab:forecasting_zero_shot} to \ref{tab:classfication_few_shot}, TimeSeek achieves state-of-the-art results across time series forecasting, anomaly detection, and classification tasks. Notably, TimeSeek requires only a small amount of downstream data for fine-tuning or even performs inference directly without any downstream training, while significantly outperforming the second-best model trained on the full downstream data across multiple benchmarks. These indicate that TimeSeek effectively learns generalized knowledge through multi-task and multi-domain pre-training, substantially enhancing its generalization and adaptability across diverse scenarios.

\vspace{-1em}
\subsection{Ablation Study}
 
\subsubsection{Multi-scale modeling} 
\begin{figure}[h]
    \centering
\includegraphics[width=1\linewidth]{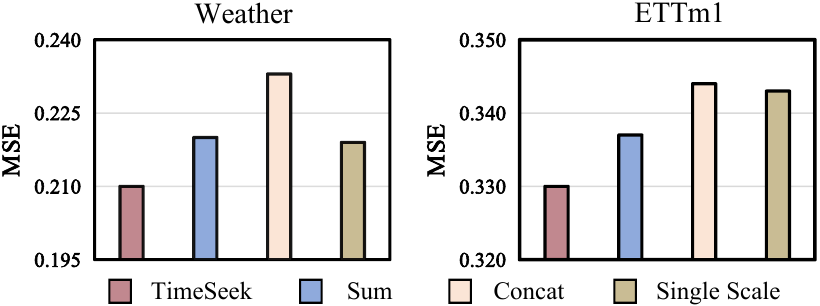}
\caption{Ablation about multi-scale modeling on Weather and ETTm1 datasets. Sum and Concat represent scale feature summation and scale feature concatenation, respectively. Single Scale denotes modeling with only the fine scale. The results are averaged across four prediction lengths.}
\label{figs:multi-scale abaltion}
\end{figure}

To evaluate the effectiveness of progressive multi-scale modeling, we conduct ablation and substitution experiments using 10\% few-shot learning. Scale Feature Summation and Scale Feature Concatenation represent two different approaches for integrating multi-scale features by addition and concatenation operations respectively to examine the necessity of adaptive fusion. Single Scale refers to using only a single fine scale for modeling, serving as a baseline to assess the necessity of multi-scale modeling.  As shown in Figure \ref{figs:multi-scale abaltion}, compared with single-scale modeling, TimeSeek achieves better prediction performance, demonstrating the effectiveness of its multi-scale modeling approach. Additionally, compared with scale feature summation and feature concatenation, TimeSeek significantly outperforms these alternative multi-scale fusion methods. Notably, Scale Feature Concatenation yields the worst prediction results, indicating that directly fusing all scale features may introduce scale feature redundancy or noise, which negatively impacts the model. This highlights the necessity of progressive and adaptive multi-scale fusion in multi-domain pre-training.

\subsubsection{Channel modeling}
\begin{figure}[h]
    \centering
    \vspace{0em}
\includegraphics[width=1\linewidth]{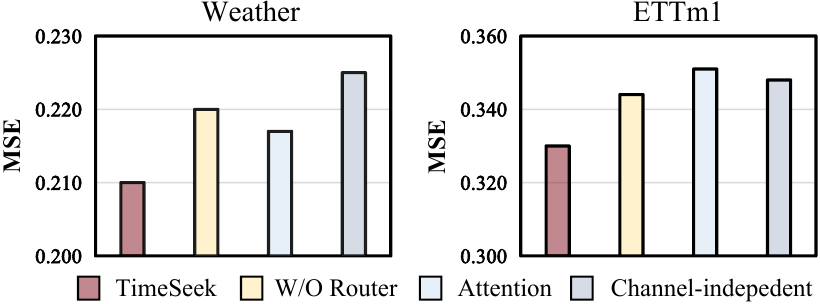}
\caption{
Ablation about channel modeling for the Weather and ETTm1 datasets. W/O Router refers to removing the channel router. Attention represents using standard attention for channel modeling. Channel-independent denotes ignoring channel modeling entirely. The results are averaged across four prediction lengths.}
\label{figs:slot attention abaltion}
\end{figure}

To validate the effectiveness of adaptive slot channel modeling, we conduct ablation and substitution experiments using 10\% few-shot learning. W/O Router refers to removing the channel router, forcing all data to undergo channel slot attention modeling. Attention replaces the proposed channel modeling with standard attention for channel processing. Channel-independent removes channel modeling entirely, ignoring inter-channel correlations.  As shown in Figure \ref{figs:slot attention abaltion}, removing the channel router leads to a drop in prediction accuracy, indicating that not all time series samples exhibit strong channel correlations. This highlights the necessity of assessing whether correlation modeling is needed for each sample. Compared to Channel-independent, TimeSeek significantly outperforms it, demonstrating that ignoring inter-channel dependencies negatively impacts performance. Additionally, compared to using standard attention for modeling channels, TimeSeek achieves better results. Notably, on the ETTm1 dataset, the standard attention-based method performs the worst, suggesting that a uniform channel modeling strategy may introduce spurious correlations between uncorrelated channels, ultimately leading to degraded performance. Therefore, an adaptive channel modeling approach is necessary to dynamically cluster channels, ensuring that strongly correlated channels are grouped together.

\subsubsection{Classification module}
To validate the effectiveness of the proposed classification module, we conduct replacement experiments on five benchmark datasets: EthanolConcentration, Heartbeat, SelfRegulationSCP1, SelfRegulationSCP2, and UwaveGestureLibrary. Our comparative analysis focused on contrasting TimeSeek's classification results with the vanilla class token approach in ViT \cite{ViT} --- a widely adopted classification method that directly initializes trainable class tokens. The experimental results demonstrate that TimeSeek achieves significant improvements in classification accuracy, outperforming the baseline method by an average margin of 17\%. This substantial performance gain indicates that TimeSeek effectively captures comprehensive global information from time series through its innovative architecture design, thereby enabling more informed classification decisions.

\begin{figure}[h]
    \centering
\includegraphics[width=0.8\linewidth]{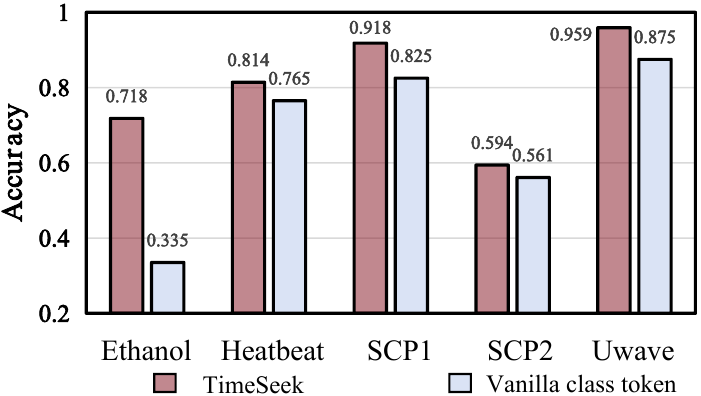}
\caption{Ablation study on the classification module across five datasets. Vanilla class token denotes directly initializing a trainable class token for classification.}
\label{figs:class token abaltion}
\end{figure}

\subsection{Model Analysis}

\subsubsection{Fine-tune with downstream data}
\begin{figure}[h]
    \centering
\includegraphics[width=1\linewidth]{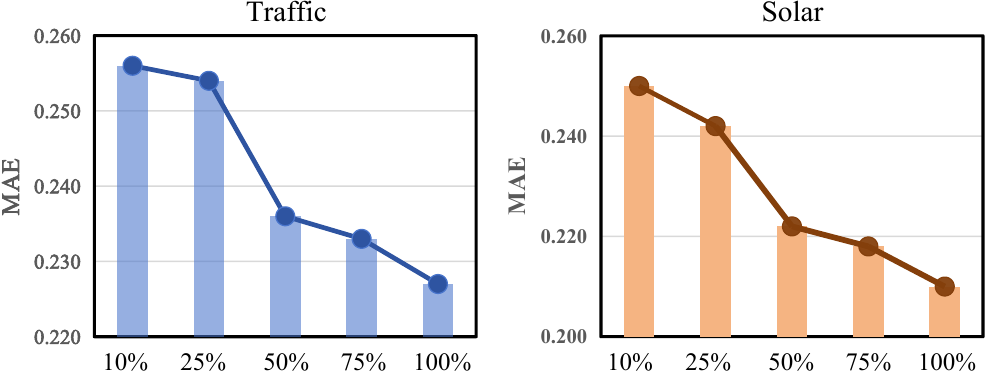}
\caption{Results of fine-tuning TimeSeek under different data percentages with prediction length 96.}
\label{figs:fine-tune with downstream data}
\end{figure}

To verify the impact of downstream fine-tuning data volume on model performance, we evaluate the model's scalability by gradually increasing the amount of downstream Traffic and Solar datasets. As shown in Figure \ref{figs:fine-tune with downstream data}, the forecast performance steadily improves as the downstream data increases, reaching best when fine-tuned with 100\% of the data. This demonstrates the strong scalability of our proposed model.

\subsubsection{Parameter sensitivity analysis.}
\begin{figure}[h]
    \centering
\includegraphics[width=1\linewidth]{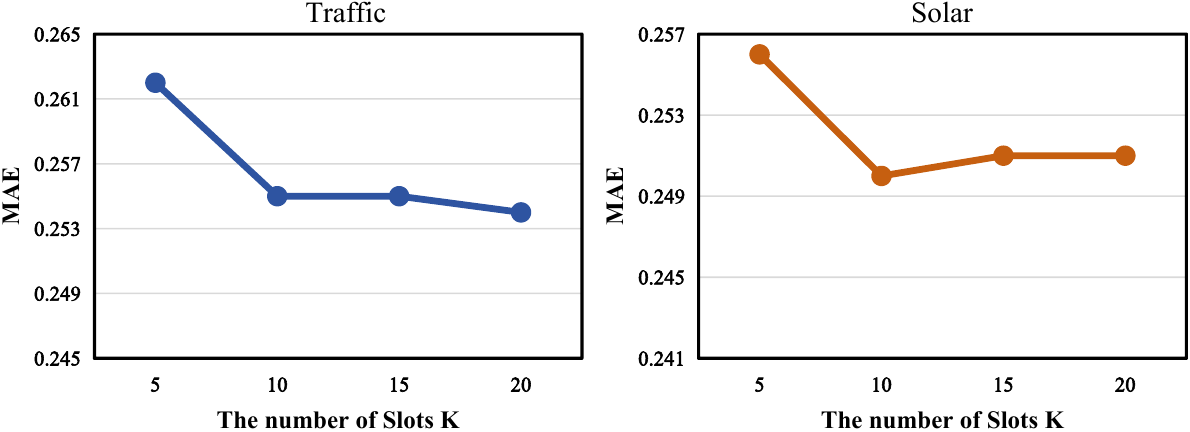}
\caption{The effect of the number of slots $K$.}
\label{figs:number_slots}
\end{figure}

We conduct parameter sensitivity experiments on Traffic and Solar datasets to investigate the effect of the number of slots $K$. Under the setting with prediction length 96, Figure \ref{figs:number_slots} reveals that insufficient slots ($K$=5) lead to suboptimal performance, indicating that a sufficient number of slots are needed to store the complex inter-channel correlations for multi-domain data. And $K$ depends on pre-training data, it may be small if data are more homogeneous. As the number of slots increases ($K$ $\leq$ 10), the model's prediction accuracy stabilizes, showing that it effectively masks redundant information while preserving key content, thus avoiding interference from irrelevant information.

\subsubsection{Efficiency analysis.}
\begin{figure}[h]
    \centering
\includegraphics[width=1\linewidth]{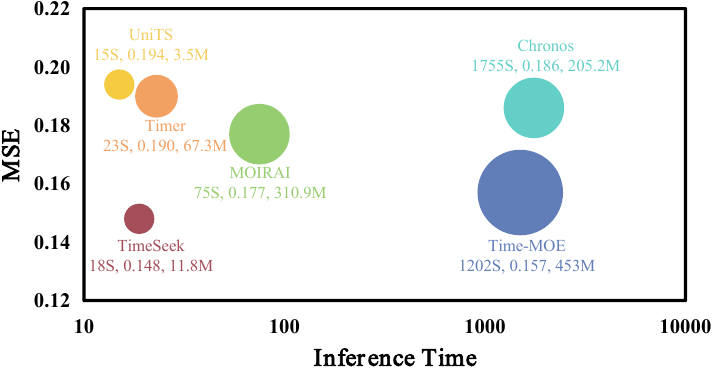}
\caption{A comparison of inference time and the parameter counts of time series foundation models, along with their zero-shot performance on the Weather dataset.}
\label{figs:effiency}
\end{figure}

Model efficiency is a key factor in determining whether a foundation model can be effectively deployed in downstream scenarios. To evaluate the efficiency of TimeSeek and other foundation models, we conduct zero-shot forecasting on the Weather dataset with a prediction length of 96. The comparison is made from three aspects: zero-shot inference time, parameter count, and prediction accuracy.  
As shown in Figure \ref{figs:effiency}, TimeSeek demonstrates significant advantages in all three aspects, highlighting its superior computational efficiency and strong applicability to downstream scenarios. Notably, compared to large-parameter models such as Time-MOE, TimeSeek achieves better inference speed, lower parameters, and higher prediction accuracy. At the same time, the pretraining cost of TimeSeek is much lower than TimeMOE. This indicates that simply increasing model size is not the only viable approach to building foundation models. Instead, effectively leveraging multi-domain time series data and diverse tasks provides a more efficient way to construct foundation models.

\section{Conclusion}
In this paper, we propose TimeSeek, a novel generalized time series analytics model that adaptively models time series patterns from multi-domain data and unifies multiple tasks in pre-training.
From the perspective of multi-domain data, we propose a Multi-Domain Feature Extractor that adaptively captures temporal dependencies and channel correlations based on domain-specific characteristics, including progressive multi-scale modeling and adaptive slot channel modeling. From the perspective of the diversity of tasks, we propose Multi-Task Collaborative Training, which leverages task specificity and task commonality through task customization and cross-task training, effectively unifying multiple tasks in pre-training and better adapting to diverse downstream scenarios. After pre-training on large-scale time series data, TimeSeek achieves state-of-the-art performance in forecasting, anomaly detection, and classification tasks across both few-shot learning and zero-shot inference situations, demonstrating strong task versatility and generalization ability. At the same time, compared with existing time series foundation models, TimeSeek shows better inference efficiency and analysis accuracy. We provide our code at https://github.com/decisionintelligence/TimeSeek.

\section{Acknowledgment}
This work was supported by the National Natural Science Foundation of China (62406112, 62372179).

\bibliographystyle{IEEEtran}
\bibliography{sample}

\vfill

\end{document}